\documentclass[letterpaper, 10 pt, conference]{ieeeconf}  

\IEEEoverridecommandlockouts                              

\usepackage{cite}
\usepackage{comment}
\usepackage{booktabs}
\usepackage{makecell}
\usepackage{graphicx}
\usepackage{tabularx,booktabs}
\usepackage{dingbat}
\usepackage{diagbox}
\usepackage[english]{babel}
\usepackage{url}
\usepackage{hyperref}
\usepackage{flushend}
\usepackage{tikz}
\usepackage{amssymb}
\usepackage{graphicx}
\usepackage{amsmath}

\title{\LARGE \bf
Using Petri Nets for Context-Adaptive Robot Explanations 
 } 

\author{G\"orkem K{\i}l{\i}n\c{c} Soylu$^{1}$, Neziha Akalin$^{1}$ and Maria Riveiro$^{1}$
\thanks{$^{1}$ Department of Computer Science and Informatics, School of Engineering,
J\"onk\"oping University, J\"onk\"oping, Sweden 
        {\tt\small neziha.akalin@ju.se, gorkem.kilinc.soylu@ju.se, maria.riveiro@ju.se}}%
}

\begin{document}

\maketitle
\thispagestyle{empty}
\pagestyle{empty}

\begin{abstract}
In human-robot interaction, robots must communicate in a natural and transparent manner to foster trust, which requires adapting their communication to the context. In this paper, we propose using Petri nets (PNs) to model contextual information for adaptive robot explanations. PNs provide a formal, graphical method for representing concurrent actions, causal dependencies, and system states, making them suitable for analyzing dynamic interactions between humans and robots. We demonstrate this approach through a scenario involving a robot that provides explanations based on contextual cues such as user attention and presence. Model analysis confirms key properties, including deadlock-freeness, context-sensitive reachability, boundedness, and liveness, showing the robustness and flexibility of PNs for designing and verifying context-adaptive explanations in human-robot interactions.

\end{abstract}

\section{INTRODUCTION}
In human-human interactions, effective communication inherently involves adapting messages based on environmental cues, situational factors, and interpersonal dynamics. Similarly, for communication between humans and AI-based systems or robots to be effective and meaningful, it must be context-sensitive. 
Since Schilit and Theimer \cite{schilit1994context} first introduced the term context-aware computing in 1994, numerous definitions of context and context-awareness have been proposed \cite{sharif2018context}. Elements used to describe contextual information are commonly grouped into five categories: individuality, activity, location, time, and relations \cite{dey2001understanding}. Although adapting a system's behavior based on detected environmental changes is not a new idea, applying this concept to explanations is underexplored. Explanations provided by robots can help users develop a better understanding of the robot's capabilities and limitations, clarify unusual behaviors or errors, and contribute to building more accurate mental models of the robot's functionality. 
It has been shown that robot explanations that clarify the reasons behind robot actions could be an effective way to avoid decreases in the trust and perceived trustworthiness \cite{edmonds2019tale, lyons2023explanations}.

Insights from context-aware computing to build context-adaptive robot explanations are highly relevant, as robots must clearly and understandably communicate their behaviors to others in the environment. Just like human explanations are naturally shaped by situational factors, robots should also be capable of adapting their explanations to fit the given context. 

This paper aims to incorporate contextual elements into robot explanations using PNs. 
PNs are a widely used formalism for modeling systems characterized by concurrency, synchronization, and causal dependencies. Originally introduced by Carl Adam Petri in the 1960s, PNs combine a precise mathematical foundation with an intuitive graphical representation, making them particularly effective in describing dynamic and distributed behavior \cite{Petri62,peterson1977petri}. 
One of the key strengths of PNs is their ability to represent concurrency, conflicts, causal dependency, and synchronization naturally in a unified platform. Places represent conditions or resources, whereas transitions model events or actions. Tokens flow through the net dynamically indicating state of the system. This structure enables modeling of systems where multiple processes occur simultaneously and may interact or compete for shared resources.

The well-defined semantics of PNs allow for the analysis of important behavioral properties such as liveness, deadlock-freeness, reachability, and boundedness \cite{murata89petri}. In addition to structural and state-based analysis, PNs support model-checking techniques in which expected properties are specified via temporal logic formulas and verified on the model. This makes PNs a powerful tool for modeling and verifying the correctness and reliability of dynamic systems.  Moreover, PNs support modular and hierarchical modeling through techniques such as abstraction, refinement, and compositionality. These methods allow large systems to be constructed from smaller, understandable components, which is particularly beneficial for complex applications such as human-robot interaction (HRI) or context-aware computing. Thus, in this paper, we propose and exemplify the use of PNs for modeling context to provide adaptive explanations in HRI. 

\section{RELATED WORK}

PNs and their extensions (e.g., timed PNs, coloured PNs, and hierarchical PNs) have been used for designing effecient multi-robot collaboration and coordination strategies. For example, Casalino et al. \cite{8809746} used timed PNs to minimize human and robot idle time in an assembly line with two robots and a human operator and showed that their method increased the productivity in the assembly line. Similarly, Chao et al. \cite{chao2016timed} developed a framework using timed PNs to model and manage multimodal behaviors such as speech, gaze, gesture and manipulation for robots. They tested their framework in a collaborative assembly task and showed that it enabled effective turn-taking and shared resource coordination during human-robot collaboration. An error recovery procedure using PNs was presented in \cite{chang2013robot}. Such procedure enabled the Nao robot to recover from both known and unknown errors by adapting task sequences or learning new recovery strategies from human demonstrations by extending the existing PN. In broader settings, Yagoda et al. 
\cite{yagoda2009modeling} used PNs for modeling human performance in a dynamic decision environment with a three-person team running unmanned aerial vehicles in search-and-rescue operations, and Figat et al. \cite{figat2019methodology} presented a methodology for designing multi-layer hierarchical PNs to model the activities of multi-agent robot systems.



Coloured PNs are used to represent task specifications and track the progress of human-robot collaboration in shared workspaces in \cite{hollerich2021coloured}. The PN framework allows the system to model and monitor the current state of tasks, incorporating partial observability and uncertainty through the novel concept of ``emissions.'' By updating the PN based on visual observations of human actions, the system can efficiently infer which actions have been performed and what steps remain, supporting dynamic negotiation and coordination between human and robot. In \cite{marrella2018measuring}, PNs are used as interaction models to represent the expected ways of performing system tasks. User actions are replayed over these PNs models to measure deviations and assess how learnable the system is during real use.


Prior research using PNs has primarily focused on collaborative tasks with industrial robots, our work  distinguishes by focusing on socially assistive robots in home environments, specifically on generating contextually adaptive explanations for trustworthy HRI. 

\section{METHOD}
In this section, we present a Petri net model of a simplified HRI scenario to demonstrate how this formalism can capture and analyze context-adaptive explanations by representing concurrent actions, conflicts, causal dependencies, local properties and changes in the system over time. One of the reasons for robot explanations is robot failures \cite{wachowiak2024taxonomy}. The scenario we model (see Section \ref{sec:examle_scenario}) is similar to the ones presented in \cite{akalin25} where a socially assistive robot in a home environment fails to complete a task and provides an explanation tailored to the context. The robot adapts its explanation modality by combining spoken explanations with an additional element (lights, error sounds, or gestures) depending on the contextual factors. We focus on a subset of context features given in \cite{dey2001understanding}, namely individuality (e.g., user attention) and location (e.g., position of the user or the robot). 


\subsection{Petri Nets}

An (ordinary) Petri net $N = (P, T, F, m_0)$ consists of a net $(P, T, F)$, and an initial marking $m_0: P \rightarrow \mathbb{N}$, where $P$ and $T$ are disjoint sets, and $F \subseteq (P \times T) \cup (T \times P)$ is called the \emph{flow relation}.
The pre-set of an element $x \in P \cup T$ is the set
${^\bullet x} = \{ y \in P \cup T \mid (y,x) \in F \}$.
The post-set of $x$ is the set
$x^\bullet  = \{ y \in P \cup T \mid (x,y) \in F \}$.
The elements of $P$ are called \emph{places}, the elements of $T$ are called
\emph{transitions}.
A net is finite if the sets of places and of transitions are finite.

\begin{figure}[h]
\centering
\includegraphics[width=7.0 cm]{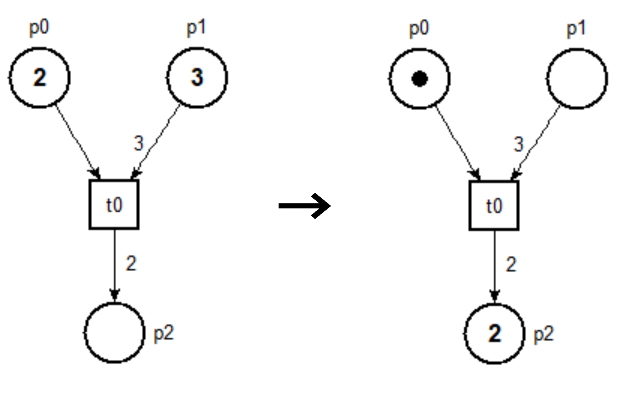}
 \caption{Firing rule with weighted arcs.}
\label{ex1}
\end{figure}

A \emph{marking} is a map $m:P \rightarrow \mathbb{N}$, which can be represented as a multi-set. For example, at the initial marking of net in Fig.~\ref{ex1} there are $2$ tokens in place $p0$ and $3$ tokens in place $p1$. There are no tokens in the other places. This marking can be denoted as $m_0 = \{2\;p0, 3\;p1\}$. 
Markings represent global states of a Petri net.

Let \( W: (P \times T) \cup (T \times P) \rightarrow \mathbb{N} \) be the arc weight function, and let \( R \subseteq P \times T \) denote the set of \emph{read arcs}. A read arc from place \( p \) to transition \( t \) checks for the presence of tokens in \( p \) without consuming them. A transition \( t \in T \) is \emph{enabled} at a marking \( m \) if, for every place $p \; \in {^\bullet}t$, either connected with an ordinary input arc or a read arc, the condition \( m(p) \geq W(p, t) \) holds. While the enabling condition is the same for both types of arcs, input arcs consume tokens upon firing, whereas read arcs do not.

If \( t \) is enabled at \( m \), then its \emph{firing} produces a new marking \( m' \), defined for all \( p \in P \) as:

\[
  m'(p) = m(p) - W(p, t) + W(t, p),
\]

where \( W(p, t) = 0 \) if there is no arc from \( p \) to \( t \), and  \( W(p, t) = 0 \) if there is no arc from \( t \) to \( p \). Read arcs do not affect the marking (i.e., tokens in places used for read arcs are preserved).
We will write $m[t\rangle m'$ to mean that $t$ is enabled at $m$,
and that firing $t$ at $m$ produces $m'$. 

In Fig.~\ref{ex1}, the illustrated PN has weighted arcs. Transition $t0$ is enabled at the initial marking since the marking satisfies the firing conditions. When $t0$ fires, it consumes 1 token from $p0$ and 3 tokens from $p1$. The number $3$ on the arc from $p1$ to $t0$ states the number of tokens to be consumed. If the number of tokens is $1$ then the number is not explicitly written on the arc. Firing of $t0$ changes the marking from $m_0= \{2\;p0, 3\;p1\}$ to $m_1= \{p0, 2\; p2\}$. The number of tokens put into a place is explicitly written on the arc if the number is greater than $1$. We write $m_0[t0\rangle m_1$ to denote that $t0$ is enabled at $m_0$ and its firing leads to $m_1$.

\begin{figure}[!ht]
\centering
\includegraphics[width=6.0 cm]{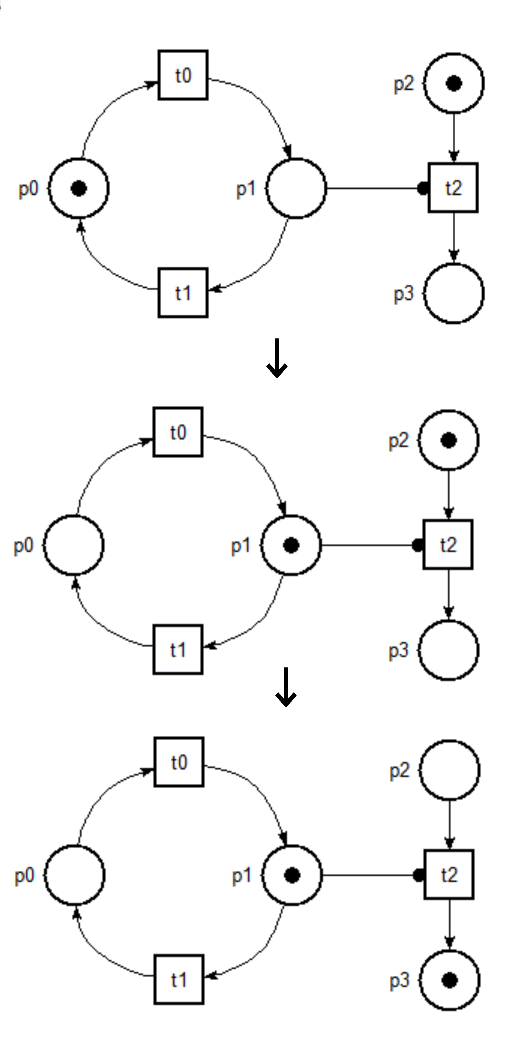}
 \caption{A Petri net with a read arc.}
\label{ex2}
\end{figure}

Fig.~\ref{ex2}  illustrates a PN with a read arc and the resulting marking changes after a sequence of transition firings.  The initial marking $m_0=\{p0,p2\}$ enables transition $t0$. After firing of $t0$, marking changes to $m_1= \{p1, p2\}$ which enables both $t1$ and $t2$. In this example, $t2$ is connected to place $p1$ via a read arc, indicating that $t2$ requires the presence of a token in $p1$ to be enabled, but does not consume that token upon firing. The last subfigure shows the marking reached after $t2$ fires.

A marking $q$ is \emph{reachable} from a marking $m$ if there exist
transitions $t_1 \ldots t_{k+1}$ and intermediate markings $m_1 \ldots m_k$ such that
\[
    m[t_1\rangle m_1[t_2\rangle m_2 \ldots m_k[t_{k+1}\rangle q
\]
%

%
A transition \( t_2 \) is said to be \emph{causally dependent} on a transition \( t_1 \) if the firing of \( t_1 \) produces tokens that are required for \( t_2 \) to become enabled. Formally, there exists a sequence of transitions such that \( t_1 \) must fire before \( t_2 \) is enabled.

Transitions \( t_1 \) and \( t_2 \) are in \emph{conflict} at marking \( m \) if they are both enabled at \( m \), but their firings are mutually exclusive due to a shared input place. That is, \( {^\bullet t_1} \cap {^\bullet t_2} \neq \emptyset \), and firing one disables the other.

Transitions \( t_1 \) and \( t_2 \) are \emph{concurrent} at marking \( m \) if both are enabled and can fire independently, i.e., they are not causally dependent, they do not share input places, and firing one does not disable the other.

\begin{figure}[!ht]
\centering
\includegraphics[width=8.0 cm]{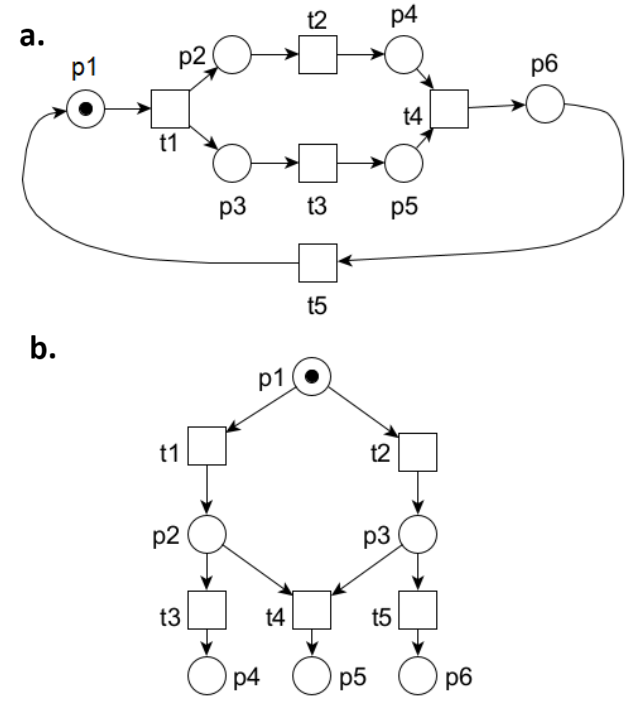}
 \caption{Two sample Petri nets.}
\label{ex3}
\end{figure}

In Fig.~\ref{ex3}, two PN systems are illustrated.  The PN in Fig.~\ref{ex3}.a is deadlock-free and live. In this net, $t1$ is enabled at the initial marking and its firing leads to marking $m_1= \{p2, p3\}$. At marking $m_1$, $t2$ and $t3$ are concurrently enabled. The two transitions can fire independently in any order. There are three possible scenarios for their firings. $t2$ can fire before $t3$, $t3$ can fire before $t2$ and they can fire simultaneously. Thus, the markings $\{p2, p5\}$, $\{p3, p4\}$ and $\{p4, p5\}$ are reachable by firing of $t2$ and $t3$. Transition $t4$ models a synchronization. It is only enabled after both concurrent transitions fire. In this net, we also see a causal dependency relation. For example, both $t2$ and $t3$ are causally dependent on transition $t1$.  Transition $t4$ is causally dependent on $t2$, $t3$ and $t1$. Transitions $t1$, $\{t2, t3\}$, $t4$ and $t_5$ are partially ordered, i.e., $t2$ and $t3$ are independent so can fire in any order after $t1$ before $t4$ but the rest of the transitions are in sequential order. 

In Fig.~\ref{ex3}.b, $t1$ and $t2$ are in conflict at the initial marking. They are both enabled at the same time, but only one can fire since they share an input place, and firing one will consume the token and disable the other transition. Thus, if $t1$ fires, the new marking reached will be $\{p2\}$, whereas if $t2$ fires the new marking reached will be $\{p3\}$. Notice that transition $t4$  needs two tokens to be enabled, one from $p2$ and one from $p3$. However, these two places are never marked together since they are marked by conflicting transitions. As a result, there is no reachable marking which enables $t4$, i.e., $t4$ is a dead transition. The net also has some dead markings such as $\{p4\}$ and $\{p6\}$. These markings do not enable any transitions. After firing of $t3$ or $t5$, the net reaches a deadlock. 


\subsection{A Sample Scenario Modeled with a Petri Net}
\label{sec:examle_scenario}
\begin{figure}[!ht]
\centering
\includegraphics[width=8.5 cm]{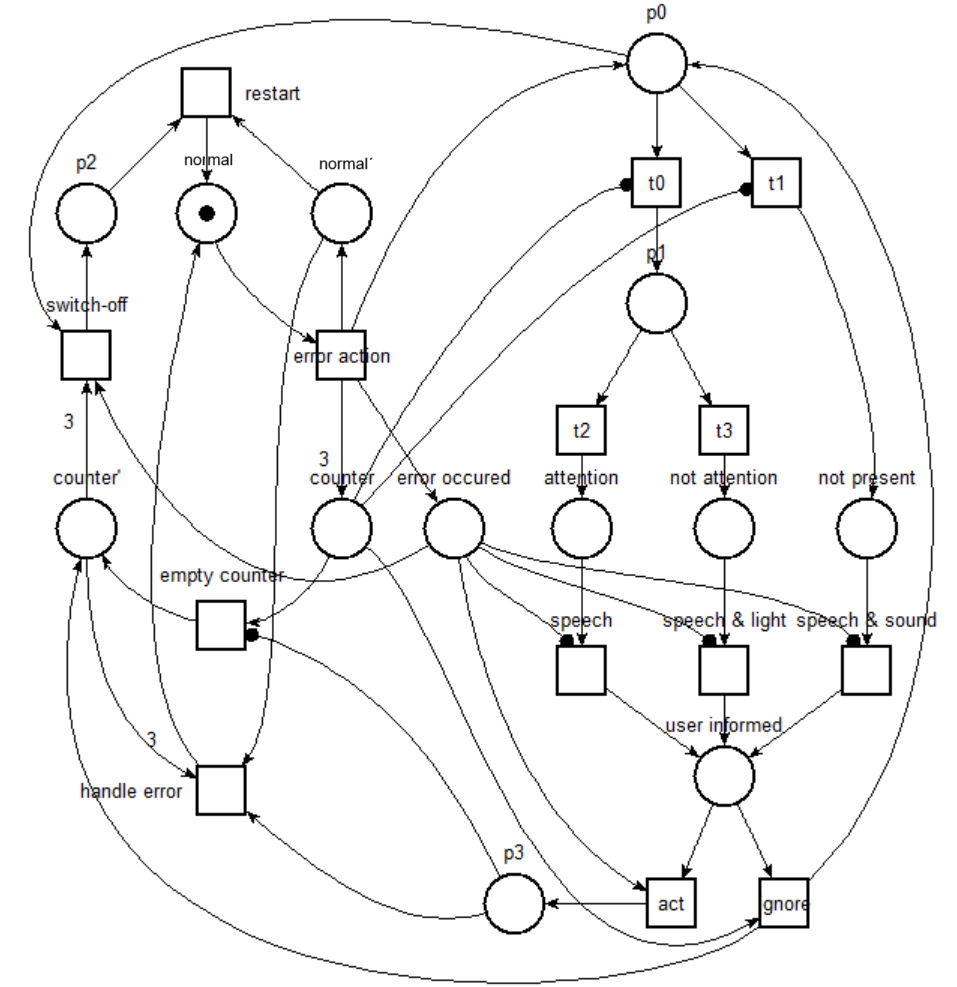}
 \caption{Petri net model of a simple scenario. }
\label{model}
\end{figure}

As previously discussed, the scenarios considered in this paper depict situations from a typical home environment with a socially assistive robot where the robot encounters an error and is unable to complete its task. These include human-robot collaboration on household tasks, shared activities such as playing games, or the robot performing chores around the home. Fig.~\ref{model} illustrates a PN model of an error-handling interaction between a robot and a human in one such scenario. In this model, initially the robot is in a normal working condition, represented by place \textit{normal} which has a token at the initial marking. The firing of transition \textit{error action} means that the robot encounters an error, which changes the state of the robot and the whole net representing the overall system consisting of the robot and the human. 
The occurrence of an error places a token in the place \textit{error occurred}, triggering the explanation process. The robot’s explanation behavior depends on two contextual factors: whether the human is present and whether they are paying attention. The place \textit{not present} represents that the human is not in the environment. The places \textit{not attention} and \textit{attention} represent two mutually exclusive states that can occur only when the human is present. At most one of these two places can hold a token at any given time, indicating whether the human is paying attention to the robot. If either \textit{attention} or \textit{not attention} is marked, it implies that the human is present and being monitored for attentiveness, while a token in \textit{not present} excludes both attention states. Depending on the marking of these places, transitions \textit{t2} and \textit{t3} guide the system toward different explanation strategies, such as using speech only, speech and light, or speech and sound. All explanation paths eventually lead to the place \textit{user informed}, indicating that the robot has attempted to communicate the error.

If the human takes an action, transition \textit{act} fires, allowing the system to return to a resolved state. Transition \textit{ignore} represents that the human does not take any action for the error, ignoring the robot's explanation. In this case, the robot continues monitoring the context. A counter mechanism is also included to track how many times the user is informed before the error is resolved. After three explanations, which can be done in different formats since the context change is monitored, the robot moves to a shutdown state, firing the \textit{switch-off} transition. This results in a state where the robot is idle, represented via place \textit{p2}. From this state, the robot can restart and re-enter its normal working condition through the firing of the \textit{restart} transition, which places a token back in the \textit{normal} place. 

\subsection{Analysis and Verification of the Model}
We analyzed the PN model to verify that it satisfies several key behavioral properties relevant to safe and adaptive HRI. These properties were confirmed through a combination of structural analysis and dynamic validation based on token flow behavior, using tools such as TINA~\cite{tina06}, PIPE2~\cite{pipe2}, and LoLA~\cite{lola2}, all of which support structural verification, reachability analysis, and model checking based on temporal logic properties over system executions.

Firstly, the model is \textit{deadlock-free}, meaning that at every reachable marking, there exists at least one enabled transition, ensuring that the system never enters a state from which no further behavior is possible. This follows from the structure of the net, even in terminal cases such as the robot shutting down or returning to the normal state, there is always a transition that is enabled, such as \textit{restart}. 

We verified several key \textit{reachability properties} that reflect correct interaction flow. The idle state (the marking $\{p2\}$), which represents shutdown, is reachable; in fact, it is guaranteed after three unsuccessful explanation attempts. If the explanation is ignored three times, the bounded counter mechanism enables the \textit{switch-off} transition once the count reaches three; at that point, it becomes the only enabled transition, forcing the system to shut down. From the resulting idle state, recovery is guaranteed since the \textit{restart} transition becomes enabled and restores the marking of \textit{normal}, confirming that the system can resume operation after termination. Together, these behaviors ensure both safe shutdown and reliable restartability.

The model also exhibits context-sensitive reachability for explanation strategies. Depending on the current marking of the context places, i.e., \textit{attention}, \textit{not attention}, and \textit{not present}, different transitions are enabled, leading to alternative explanation modalities (\textit{speech}, \textit{speech \& light}, \textit{speech \& sound}). For example, \textit{speech \& light} is only enabled when \textit{attention} is marked and \textit{not present} is unmarked. We confirmed that these context-driven branches are reachable and mutually exclusive.

Another important dynamic property is the \textit{boundedness} of the explanation counter. After three explanation cycles, tokens accumulate in the \textit{counter'} place, enabling the shutdown behavior. We verified that this value never exceeds the bound of three by construction. The net enforces token flow through places \textit{counter} and \textit{counter'} in such a way that no unbounded accumulation occurs. This was validated through invariant checking and coverability analysis.

The model also satisfies \textit{liveness} for all critical transitions involved in the interaction. Once the robot reaches the \textit{user informed} state, both the \textit{act} and \textit{ignore} transitions become enabled, reflecting the possibility for the human either to respond or to disregard the explanation. Depending on which transition fires, the system either resolves the error and returns to a stable state or continues with additional explanation attempts, potentially leading to shutdown. These transitions remain live in all relevant reachable markings, and their liveness was confirmed through temporal logic queries.

In addition to the properties discussed above, the model satisfies a basic \textit{fairness property}. Whenever the place \textit{user informed} is marked, representing the state in which the user is informed, the transition \textit{act} gets enabled. This means that the human is always given an opportunity to resolve the error before the robot shuts down.

Finally, the model satisfies a crucial \textit{safety property} that prevents premature shutdown. Specifically, the robot is not allowed to enter the idle state (represented by marking $\{p2\}$), unless it has first attempted to explain the error three times. This is guaranteed by unreachability of any marking in which place $p2$ is marked while fewer than three tokens have passed through the counter mechanism. Structurally, the transition \textit{switch-off} is only enabled when the \textit{counter'} has reached its maximum value, ensuring that shutdown can only occur after three explanation cycles. 

The simplicity and modularity of the model also make it highly extensible. Additional parameters such as the criticality level of the error, time constraints, user preferences, or environmental factors could be incorporated by adding new places, transitions, or contextual conditions. Moreover, concurrent processes such as monitoring multiple users or coordinating with other robots can be integrated through subnets. Techniques such as \textit{abstraction} and \textit{refinement} can be applied to manage the complexity as the model grows. These allow verification of high-level interaction logic and then incremental refinement to include detailed behaviours or internal logic of components. This flexibility makes PNs a promising formal tool for designing and verifying HRI. 

\section{CONCLUSION}
This paper proposed employing PNs for context-aware robot explanations, especially for adapting the robot explanation modality. By incorporating contextual elements such as user presence and attention, we demonstrated how PNs can effectively represent and manage explanation behaviors that respond to dynamic interaction conditions. We showed that our PN model satisfies critical properties for safe and adaptive interaction, including deadlock-freeness, boundedness, liveness, and context-sensitive reachability. Future work will explore incorporating more relevant contextual elements for providing trustworthy, adaptive and context-aware explanations.






\section*{ACKNOWLEDGMENT}
This work was partly funded by the Swedish Research Council, project VR 2022-03180 XPECT -- How to tailor explanations from AI systems to user's expectations.


\bibliographystyle{IEEEtran} 
\bibliography{ref}
\end{document}